\documentclass{article}

    \PassOptionsToPackage{numbers, compress}{natbib}

    \usepackage[preprint]{nips23}

\usepackage[utf8]{inputenc} 
\usepackage[T1]{fontenc}    
\usepackage{hyperref}       
\usepackage{url}            
\usepackage{booktabs}       
\usepackage{nicefrac}       
\usepackage{microtype}      
\usepackage{xcolor}         

\usepackage{bookmark}

\usepackage{graphicx}

\usepackage{multirow}
\usepackage{multicol}
\usepackage{amsmath,amsfonts,amsthm,amssymb,amsopn,bm}
\usepackage{xspace}
\usepackage{tcolorbox}
\usepackage[ruled,linesnumbered]{algorithm2e}

\usepackage{subcaption}

\newcommand{\F}{Fig.}
\renewcommand{\F}{Figure}
\newcommand{\T}{Table}
\renewcommand{\S}{Section}

\newcommand{\tool}{\textsc{VRPTest}\xspace}

\newcommand{\parh}[1]{\noindent\textbf{#1}}
\newcommand{\vsp}{visual referring prompting}

\newcommand{\finding}[2]{
  \smallskip
  \smallskip
\begin{tcolorbox}[width=\linewidth,boxrule=0pt,top=1pt, bottom=1pt, left=1pt,right=1pt, colback=gray!20,colframe=gray!20]
\textbf{Finding #1:} 
{#2}
\end{tcolorbox}}

\title{VRPTEST: Evaluating Visual Referring Prompting in Large Multimodal Models}

\author{Zongjie Li\textsuperscript{1}, Chaozheng Wang\textsuperscript{2}, Chaowei Liu\textsuperscript{4}, Pingchuan Ma\textsuperscript{1} \\ 
\textbf{Daoyuan Wu\textsuperscript{3}\textsuperscript{*}, Shuai Wang\textsuperscript{1}\thanks{Corresponding authors.} , Cuiyun Gao\textsuperscript{2}}\\
\textsuperscript{1}Hong Kong University of Science and Technology,\textsuperscript{2}Harbin Institute of Technology\\
\textsuperscript{3}Nanyang Technological University, \textsuperscript{4}National University of Singapore \\
\texttt{\{zligo,pmaab,shuaiw\}@cse.ust.hk, daoyuan.wu@ntu.edu.sg}, \\
\texttt{wangchaozheng@stu.hit.edu.cn, gaocuiyun@hit.edu.cn},
\texttt{e1011116@u.nus.edu},
}

\begin{document}

\maketitle

\begin{abstract}
    With recent advancements in Large Multimodal Models (LMMs) across various domains, a novel prompting method called \textit{visual referring prompting} has emerged, showing significant potential in enhancing human-computer interaction within multimodal systems.
    This method offers a more natural and flexible approach to human interaction with these systems compared to traditional text descriptions or coordinates.
    However, the categorization of visual referring prompting remains undefined, and its impact on the performance of LMMs has yet to be formally examined.
    In this study, we conduct the first comprehensive analysis of LMMs using a variety of visual referring prompting strategies.
    We introduce a benchmark dataset called \tool, comprising 3 different visual tasks and 2,275 images, spanning diverse combinations of prompt strategies.

    Using \tool, we conduct a comprehensive evaluation of eight versions of
    prominent open-source and proprietary foundation models, including two early
    versions of GPT-4V. 
    We develop an automated assessment framework based on software metamorphic testing techniques to evaluate the accuracy of LMMs without the need for human intervention or manual labeling.
    We find that the current proprietary models generally
    outperform the open-source ones, showing an average accuracy improvement of
    22.70\%; however, there is still potential for improvement. Moreover, our
    quantitative analysis shows that the choice of prompt strategy significantly
    affects the accuracy of LMMs, with variations ranging from -17.5\% to
    +7.3\%. Further case studies indicate that an appropriate \vsp\ strategy can
    improve LMMs' understanding of context and location information, while an
    unsuitable one might lead to answer rejection. We also provide insights on
    minimizing the negative impact of \vsp\ on LMMs. This study aims to set a
    benchmark and guide future research in \vsp.
    The benchmark and the metamorphic-based framework developed for
    generating datasets are available at
    \texttt{https://github.com/tszdanger/VisualReferPrompt}.

\end{abstract}

\section{Introduction}
\label{sec:intro}

Recent advances in large language models (LLMs) have achieved remarkable performance on various tasks, sometimes even surpassing human capabilities~\cite{kojima2022large,thapa2023humans,wang2023reef}. Inspired by the success of LLMs, researchers have proposed large multimodal models (LMMs) to extend LLMs to multimodal scenarios involving both language and visual inputs. In contrast to traditional vision systems where inputs and outputs follow fixed formats, LMMs can process arbitrarily interleaved multimodal inputs and outputs. 
While mature LMMs comparable to LLMs like ChatGPT have yet to emerge, preliminary LMMs like GPT-4V~\cite{yang2023dawn} have demonstrated promise on a diverse range of tasks. These include conventional vision tasks like image classification~\cite{lu2007survey}, generation~\cite{wu2017survey}, and object detection~\cite{zhao2019object}.
Given their versatile capabilities, LMMs have shown potential for numerous real-world use cases, including in the medical domain~\cite{singh2021medical}, GUI navigation~\cite{wang2023unified}, and so on.

Pointing to specific spatial locations is an essential capability for human-computer interaction with multimodal systems, such as conducting visually grounded dialogues. Instead of using traditional text descriptions or coordinates, a natural idea is to directly edit the image pixel space to draw visual pointers or scene texts as human referring instructions. This gives rise to a new prompting method called \textit{visual referring prompting} (VRP).
Generally, \vsp\ is a wide range of prompting methods that use visual pointers or scene texts to guide the multimodal systems to focus on a specific spatial location in the image, and further help the multimodal systems to generate the corresponding outputs.

\begin{figure}[!htpb]
    \centering
    \includegraphics[width=1.0\textwidth]{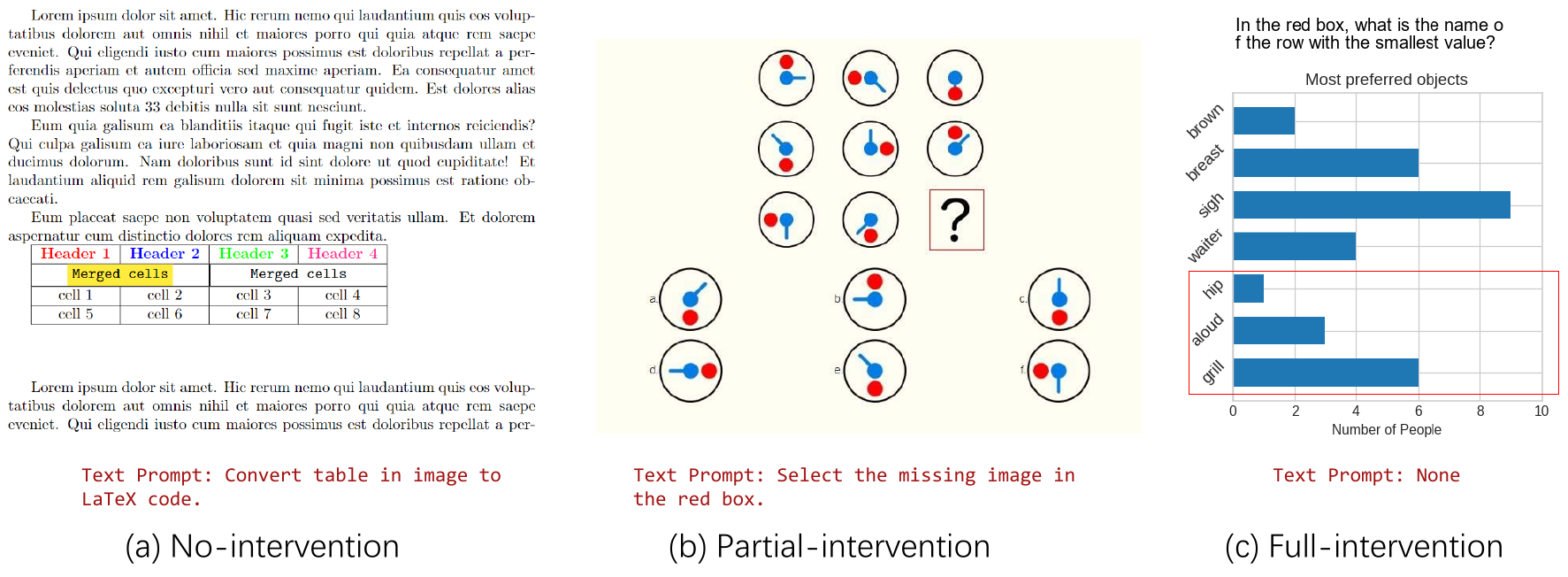} 
    \caption{A conceptual diagram of three types of \vsp. \textbf{Left}: No-intervention, \textbf{Middle}: Partial-intervention, \textbf{Right}: Full-intervention.}
    \label{fig:example}
\end{figure}

Based on the extent of human intervention entailed, \vsp\ can be categorized into three types: \textit{No-intervention}, \textit{Partial-intervention}, and \textit{Full-intervention}, as shown in~\F~\ref{fig:example}. 
As the name implies, \textit{No-intervention} does not provide any visual referring prompts to the multimodal system. The instructions rely solely on a text prompt, typically containing text coordinates or descriptions of the target location.
In \textit{Partial-intervention} VRP, the multimodal system receives a visual pointer (e.g., a bounding box) to guide focus toward a specific spatial location in the image. The primary instructions still come from the text prompt, but the target location is more precisely identified in the image. 
In contrast, \textit{Full-intervention} VRP does not require a text prompt. It merges the text and visual prompts into a single image, with instructions patched directly onto the image and the target location highlighted by a visual pointer.

Numerous datasets focusing on various visual tasks have been created to assess the performance of LMMs, and it has been shown that LMMs can achieve remarkable performance on these traditional tasks such as visual question answering (VQA)~\cite{antol2015vqa} and visual dialog (VD)~\cite{meng2020openvidial}. 
Recently, new benchmarks have been proposed to further evaluate the advanced capabilities of LMMs, such as mathematical reasoning ability~\cite{lu2023mathvista} and multi-step instruction following ability~\cite{guo2023pptc}.
However, the impact of \vsp\ on the performance of LMMs has not been formally examined, 
and there is a lack of a benchmark dataset as well as a testbed for \vsp\ in LMMs.

In this work, we introduce \tool, a consolidated \underline{\textbf{V}}isual
\underline{\textbf{R}}eferring \underline{\textbf{P}}rompting benchmark
containing 2,275 image-question pairs across three distinct visual tasks and 12
different prompt strategy combinations. To construct this benchmark, we first
meticulously collect images and craft corresponding questions amenable
to VRP. Then, we leverage software metamorphic testing (MT)~\cite{chen1998metamorphic} to systematically generate new prompt strategy combinations by mutating the original image/question pairs following a set of
well-designed metamorphic relations (MRs). MT alleviates the need for manual
labeling and human intervention, enabling automated evaluation of LMMs using our
created image/question inputs and their MR-mutated counterparts.

Our study reveals several key insights into the effects of \vsp\ on LMMs: 
1) The average accuracy of proprietary models is 22.70\% higher than that of open-source models, but there is still a lot of room for improvement.
2) The prompt strategy significantly impacts LMM performance, and careful design enables otherwise underperforming models to excel. 
3) GPT-4V derives greater benefits from VRP compared to
open-source models, with performance being more sensitive to the prompt
strategy. In summary, our study makes the following main contributions:

\begin{itemize}
    \item We for the first time investigate the impact of \vsp\ on large
    multimodal systems, which has proven effective at enhancing human-AI
    interaction but lacks a formal study.
    \item We introduce and release a benchmark dataset, \tool, to enable
    evaluation, along with a metamorphic-based framework for systematic
    and continuous data generation.
    \item We present a quantitative analysis of \vsp\ strategies to elucidate their effects on LMMs' performance.
    \item Several case studies are conducted to provide further findings on the impact of \vsp\ on GPT-4V.
\end{itemize}

\section{The \tool Benchmark}
\label{sec:method}

As we introduced in \S~\ref{sec:intro}, we propose a benchmark with three different tasks and the toolkits to evaluate the performance of LMMs on \vsp.
In this section, we will provide an overview of the three tasks, including the image collection or generation process, as well as the fundamental metadata for each task at different levels of difficulty. Additionally, we will introduce the central concept of our toolkits, which employs a metamorphic-based framework to generate test images with varying \vsp\ strategies. Finally, we will outline the evaluation metrics for the three tasks in \tool, which are utilized to assess the efficacy of LMMs on \vsp.

\subsection{Original Image: Collection and Basic Annotation}
\label{subsec:data-collection}

In this section, we introduce the process of data collection for the three tasks in \tool, which results in a total of 174 image-question pairs as the original images.

\parh{IQtest.}~We curate the IQtest dataset from some public IQ test websites, including CSUN~\cite{csuntest}, 123Test~\cite{test123}, and BRGHT~\cite{brghttest}.
The question format of the IQtest dataset is simple, where an image is provided and a multiple-choice question is asked based on the image. 
After collection, we manually verify and remove any original embedded prompts to preclude influence on \vsp.
Two annotators independently label each question, discussing disagreements to reach a consensus.
As the sole task with single-answer multiple choice questions, IQtest difficulty is marked as ``low''. In total, the IQtest dataset comprises 40 questions.

\parh{Reasoning.}~The reasoning dataset is mainly derived from newly collected dataset MathVista~\cite{lu2023mathvista}, which is based on 28 different source datasets.
For each image, we manually check the type and the content of the image to make sure that the image is suitable for further extension with \vsp. Specifically, the images related to portraits are excluded from the dataset to avoid potential personal information leakage.
Notably, we do not reuse the questions of MathVista, but manually create the questions based on the images that we view as more suitable for the reasoning task and can be easily represented by the \vsp.
Compared to IQtest, images in the Reasoning dataset are more complex, and the questions are more diverse in format, and thus the difficulty of the Reasoning dataset is marked as ``medium''. 
Finally, we obtained a total of 124 questions in the Reasoning dataset.

\parh{LatexTab.}~To collect the LatexTab dataset, we first collect the source
code of latex tables from various resources, including the latex
forums~\cite{latexforum} and latex tutorial websites~\cite{latextutorial}. After
collecting the source code, we do not directly use them as the questions in the
dataset, but manually make some modifications to balance the difficulty as well
as the novelty of the questions. Next, we compile the latex source code to
obtain the corresponding images. In this step, we use the \textit{pdflatex}
compiler and insert some meaningless and arbitrary text contents around the
latex table to mimic the real-world scenario, as shown in~\F~\ref{fig:example}
(a). The question format of the LatexTab dataset is fixed, where a figure of the
latex table is provided and a question is asked to give the source code of the
latex table. As it is extremely difficult to reproduce the latex code word for
word, we mark the difficulty of the LatexTab dataset as ``high''. In total, we
obtained a total of 10 questions in the LatexTab dataset.

\parh{Avoid Data Leakage.}
To preclude performance overestimation from data leakage, we implement two strategies when constructing \tool. First, all data is collected between \textbf{2023-10-10} and \textbf{2023-10-21}, after the release of all evaluated LMMs. Second, images are either manually created or sourced from public domains and then modified to alter the content. These steps ensure our test data has zero overlap with any potential LMM training corpora. As the training data of benchmarked models are proprietary, directly verifying leakage is infeasible. However, by sourcing data after deployment and altering image contents, we believe \tool\ mitigates data leakage risks and produces reliable results. Overall, the combination of temporal data collection boundaries and image content modification aims to benchmark LMMs on fully unseen data, avoiding biases from data leakage and providing rigorous VRP evaluation.

\subsection{Metamorphic-based Framework}
\label{subsec:metamorphic-extension}

\begin{table}[th!]
    \centering
    \begin{tabular}{c|c|c}
        \toprule
        \textbf{MR Name} & \textbf{Description} & \textbf{Options}\\
        \midrule
        Color & The color of the reference texts and signs. & [Red | Blue | Default] \\
        Font Type & The font type of the reference texts.& [Arial | TNR | None]\\
        Refer shape & The shape of the reference signs.& [Box | Ellipse | None]\\
        Refer Position & The position of the reference texts.& [ Upper | Lower | None]\\

        \bottomrule
    \end{tabular}
    \caption{MRs for \vsp.}
    \label{tab:MR-definition}
    \end{table}

\parh{Metamorphic Testing (MT).}~MT is a widely used technique for testing deep neural networks (DNNs) when the ground-truth answers are either explicitly defined or unknown~\cite{zhang2018deeproad,wang2020metamorphic}.
The fundamental idea behind MT is to establish a set of metamorphic relations (MRs) that describe how the input should be transformed and the expected relationship between the output and the input~\cite{chen1998metamorphic}. Each MR in MT comprises two components: a metamorphic transformation (MRt) and a relation (MRr). The MRt specifies a mutation scheme that alters the source input to produce a follow-up test input, while the associated MRr defines the expected output relationship between the source input and the mutated input.
For example, to test the trigonometric function $sin(x)$, an MR can be constructed such that its MRt mutates the input $x$ into $t-x$, and the MRr checks the equality relation $sin(x) = sin(t-x)$. 
In real-world usage, MRr typically denotes invariant program properties that should always hold when the input is arbitrarily mutated using MRt, and a bug is detected whenever MRr is violated.

\parh{Metamorphic Relations for Visual Referring Prompting.}
To enable analysis of \vsp\ strategies' impacts, we expand the dataset by
applying MRs on the original images. Our goal is to incorporate diverse interventions resembling real-world usage while keeping them effective for evaluation.
Based on analyzing examples in~\cite{yang2023dawn} and user exploration patterns, we identify and define four key MRs in~\T~\ref{tab:MR-definition}. Critically, each relation has a ``None'' option to represent no transformation for the baseline ``No-intervention'' prompting. For the ``Partial-intervention'' prompting,
transformations involving reference texts are omitted since the prompting
questions remain unchanged. Overall, the MRs systematically incorporate
additional visual and textual signals into the original images. By applying
combinations of these relations, we construct a comprehensive set of intervened
variants from the original images. The expanded dataset \tool\ enables assessing
LMM performance under diverse visual prompting strategies and composite
interventions.

\subsection{Metadata Annotation}
\label{subsec:metadata-annotation}

The final stage involves the annotation of images with metadata, which is essential for a comprehensive analysis of models' capabilities across various \vsp\ strategies. 
Overall, the metadata annotation features can be divided into two parts: basic metadata and additional metadata. 
The basic metadata, as we introduced in \S~\ref{sec:intro}, are collected during the data collection process, which includes the dataset name, image index, image name, question content, and ground truth answer. 

Moreover, as the metamorphic-based framework is applied to generate variants of the original images (see
details in~\S~\ref{subsec:metamorphic-extension}), we also annotate these variants with additional metadata. These additional metadata include color, font type, reference shape, and reference position. Notably, all the variants are generated automatically via our metamorphic-based framework, and thus no additional manual annotation is required for the additional metadata.

For better understanding, we provide a conceptual illustration of key metadata
in examples with \vsp\ in~\F~\ref{fig:source_dataset}. As illustrated, each test
image (left) contains basic metadata (orange background) along with additional
metadata (yellow background). The additional metadata consists of two
components: reference text describing the target object, and visual signs
highlighting the target in the image. Notably, not all test images contain full
additional metadata --- the example depicts the ``Full-intervention'' with both
reference text and visual signs. Other \vsp\ strategies may incorporate reduced
metadata. Overall, the schematic summarizes the metadata composition, including
basic information for all test cases as well as optional reference texts and
visual signs for \vsp.

\subsection{Dataset Analysis}
\label{subsec:dataset-analysis}

\begin{figure}[t!]
     \begin{minipage}{0.35\textwidth} 
     \centering
     \fontsize{7.5pt}{\baselineskip}\selectfont 
     \renewcommand\tabcolsep{0.9pt} 
     \renewcommand\arraystretch{0.83} 
     \begin{tabular}{lc}
     \toprule
     \textbf{Statistic} & \textbf{Number} \\
     \midrule
      Total Questions & 2,275 \\
      ~- IQtest questions & 520 (22.86\%) \\
      ~- Reasoning questions & 1,625 (71.43\%) \\
      ~- LatexTab questions & 130 (5.71\%) \\
     \midrule
     Unique number of original images & 174 \\
     Unique number of variants& 2,088 \\
     \midrule
     No-intervention & 174(7.69\%) \\
     Partial-intervention & 696(30.77\%) \\
     Full-intervention & 1392(61.54\%) \\
     \midrule
     Maximum question length & 158 \\
     Maximum answer length & 24 \\
     Average question length & 65.3 \\
     Average answer length & 4.5 \\
     \bottomrule
     \end{tabular}
     \captionof{table}{Key statistics of \tool.}
     \label{tab:statistics}
     \end{minipage} 
     \hfill
     \begin{minipage}{0.58\textwidth}
     \centering
    \includegraphics[width=1.0\linewidth]{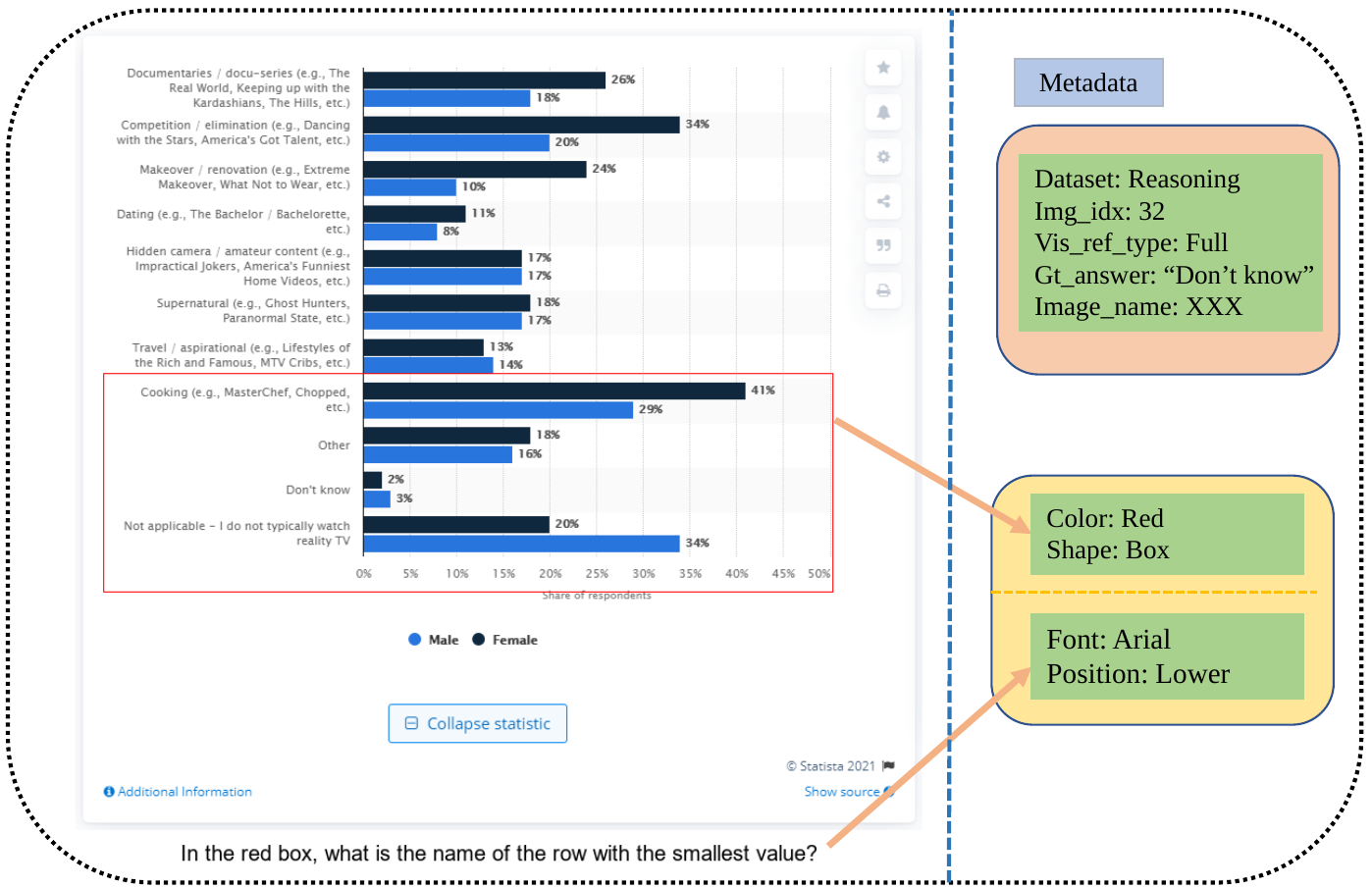}
     \caption{A conceptual illustration of key metadata in examples with \vsp.}
     \label{fig:source_dataset}
     \end{minipage}
    \end{figure}

The main statistics of \tool are presented in~\T~\ref{tab:statistics}. The
Reasoning dataset contributes the majority of questions at 71.43\%, while
LatexTab comprises the smallest portion at 5.71\%. This imbalance stems from the
difficulty of eliciting meaningful responses for high-difficulty questions
during preliminary experiments. Thus, only a small number of overly complex
questions were included. Furthermore, the full-intervention variants are the
most numerous, while No-intervention is the least. This results from the
multiplicity of \vsp\ strategy combinations applied to the base images --- far
exceeding the original seed count. 

Additionally, the average and maximum question lengths substantially exceed the answer lengths in~\T~\ref{tab:statistics}. This discrepancy stems from the focused question design eliciting clear, concise responses rather than open-ended descriptions. The questions provide full details and context to guide the desired image interpretation, while the answers concisely convey key elements. For instance, questions may describe the overall scene and components prompting identification of specific attributes or relationships. In contrast, the answer highlights the requested details succinctly.

\parh{Cost.}
The total cost of the project comprises two components, namely the human-related cost and the computing cost. The human-related cost covers the compensation paid to the annotators for their efforts in performing annotation and human evaluation tasks. Specifically, the annotators were compensated at an average rate of 5 USD per hour (refer to \S~\ref{subsec:data-collection} and \S~\ref{subsec:evaluation-protocols} for detailed information). 
On the other hand, the computing cost only applies to the proprietary models, where the API is charged on a pay-as-you-go basis. Since the open-source models are deployed locally, the computing cost is not taken into account. 
Overall, the total budget allocated for annotations and other related expenses throughout the project duration amounted to 840 USD.

\section{Experiments}
\label{sec:method}

\subsection{Evaluation Process}
\label{subsec:evaluation-protocols}

Recent LLMs and LMMs have been instructed to generate long responses in conventional settings instead of short texts, which increases the difficulty of evaluation.
Therefore, our evaluation process consists of three stages: \textit{response generation}, \textit{answer extraction}, and \textit{score calculation}. Initially, the tested LMMs generate answer texts given the input query, which incorporates the task description, the question, the choices (if applicable), and the image.
Then, we follow~\cite{li2023split} and propose an answer extractor based on ChatGLM3, inspired by its remarkable ability for text processing~\cite{chatglm2}. A preliminary study of 200 examples shows that ChatGLM3 can extract the answer text with more than 97\% accuracy. 
Finally, the extracted answer is normalized and compared with the ground truth answer to measure the performance.

\subsection{Experimental Setup}
\label{subsec:experimental-setup}

\parh{Tested Models.}
The LMMs evaluated encompass both publicly available open-source models as well
as proprietary models. The open models tested include
mPLUG-OWL~\cite{ye2023mplug}, a multimodal model provided by Alibaba;
miniGPT-4~\cite{zhu2023minigpt}, a vision-language multi-task learning model
from KAUST; InstructBLIP~\cite{instructblip}, a vision-language model with
instruction tuning; LLaVA~\cite{liu2023llava}, a model trained on LAION and VQA
datasets; and CogVLM~\cite{wang2023cogvlm}, which contains a trainable visual
expert module in the attention and FFN layers. Additionally, we evaluate the
proprietary GPT-4V~\cite{openai2023gpt4v} from OpenAI, as it offers two
distinct versions --- a web service and a paid API access. Both the web service
and API service are tested given their unique capabilities. It is worth noting
that due to the strict anti-automation testing of the GPT-4V web service, we
manually test all samples excluding the variants. In total, our experiments
assess eight different versions of LMMs spanning open and proprietary
categories. The diversity of models provides a comprehensive evaluation of the
\vsp's impact on current LMMs.

\parh{Metrics.}
As introduced in~\S~\ref{sec:method}, the IQtest and Reasoning datasets contain questions with ground truth answers. After extracting model responses, accuracy is calculated by comparing the predicted and true answers, with the total accuracy defined as the percentage of correctly answered questions out of all attempted questions. For the LatexTab datasets, given the potential for multiple mathematically equivalent answers expressing the same relationship, a strict exact match is insufficient. Therefore, we follow previous works~\cite{li2023protecting,li2022cctest} and employ a relaxed accuracy metric using BLEU-4~\cite{papineni2002bleu} scoring between the generated and ground truth answers. Responses with a BLEU-4 score exceeding a predefined threshold of 0.8 are considered correct.

\parh{Prompt Adjustment.}
During experiments, we find that the performance of LMMs is sensitive to the prompt, which is consistent with the findings in previous works~\cite{liu2023hallusionbench}.
Moreover, optimal prompts differ across LMM architectures, and using a static prompt risks unfairly disadvantaging certain models. 
Therefore, we reference official guidelines and construct up to four prompt templates per model. During evaluation, we test all versions and select the best-performing prompt individually for each LMM. This customized approach aims to elicit the full capabilities of each model architecture. The prompt formulations are detailed in~\cite{VisualReferPrompt}.

\parh{Reproducibility.}
To ensure reproducible results, sampling is disabled for locally run models to obtain deterministic outputs. However, proprietary LMMs accessed via web service lack interfaces to control hyperparameters. Therefore, we mitigate variability by executing multiple trials and aggregating representative results through a majority vote. All experiments utilize a standardized computational platform with an Intel
Xeon Platinum 8276 CPU, 256 GB of main memory, and 4 NVIDIA A100 GPUs.

\subsection{Main Results}
\label{subsec:results}

\T~\ref{tab:result-acc} compares LMMs' performance on \tool\ across the IQtest and Reasoning datasets. For each model, we report accuracy under the ``No-intervention'', ``Partial-intervention'', and ``Full-intervention'' conditions outlined in~\S~\ref{sec:intro}. The results on the LatexTab dataset are not reported here, as no example has been successfully converted for any open-source LMMs.
Overall, results reveal ample room for improvement, with all open-source LMMs underperforming commercial models like GPT-4V. On average, open-source accuracy lagged GPT-4V by 22.70\% across datasets and conditions. The sole exception was miniGPT-7B matching GPT-4V performance on IQtest under No-intervention, but still trailing on Reasoning. 

Additionally, we find that performance patterns diverge across the IQtest and Reasoning datasets. As discussed in~\S~\ref{subsec:data-collection}, IQtest employs simpler question formats perceived as less difficult. However, contrary to expectations, average open-source LMM accuracy on IQtest proved comparable to Reasoning without intervention. What's more, GPT-4V performed worse on IQtest than Reasoning in the No-intervention setting.

The impact of intervention levels on model performance varied depending on the model and dataset used. 
Partial-intervention is found to be beneficial for InstructBLIP-7B and miniGPT-7B on IQtest, where their original performance is low. 
For other models, Partial-intervention has a neutral or detrimental effect compared to No-intervention. 
On the other hand, Full-intervention results in competitive or superior accuracy for open-source LMMs and GPT-4V on both IQtest and Reasoning.
In fact, Full-intervention ranks first in 9 out of 14 \vsp\ strategy combinations, indicating its overall effectiveness. 
Numerical analysis reveals that different intervention strategies could result in up to a 17.5\% negative gain (GPT4V, partial) or a 7.3\% gain (CogVLM, full). Overall, the results suggest that the effectiveness of intervention strategies can vary depending on the model and dataset, where Full-intervention may be a more effective approach in some cases.

\finding{1}{Compared to the commercial LMM such as GPT-4V, the performance of the open-source LMMs is on average 22.70\% lower. Different levels of intervention have various effects on the performance across LMMs.}

\begin{table*}[!htbp]
    \small
    \centering
    \begin{tabular}{c|c|c|c|c|c|c}
        \toprule
        {\textbf{LMM}}         & \multicolumn{3}{c|}{IQ} & \multicolumn{3}{c}{Reasoning} \\  \midrule
                        & NO    & PARTIAL & FULL   & NO    & PARTIAL & FULL   \\ \midrule
        CogVLM        & 0.175 & 0.175   & 0.2625 & 0.224 & 0.16    & 0.223  \\ \midrule
        InstructBLIP-7B~\cite{instructblip}     & 0.125 & 0.13125 & 0.1625 & 0.264 & 0.182   & 0.139  \\ \midrule
        LLaVA-7B~\cite{liu2023llava}     & 0.2142 & 0.2121 & 0.2225 & 0.264 & 0.184   & 0.171  \\ \midrule
        LLaVA-13B~\cite{liu2023llava}     & 0.175 & 0.14375 & 0.1719 & 0.176 & 0.126   & 0.141  \\ \midrule
        mPLUG-7B~\cite{ye2023mplug}              & 0.2   & 0.175   & 0.2125 & 0.072 & 0.046   & 0.108  \\ \midrule
        miniGPT-7B~\cite{zhu2023minigpt}    & 0.175 & 0.1812  & 0.2156 & 0.264 & 0.24    & 0.28   \\ \midrule
        Average    & 0.1773 & 0.1697  &  0.2079  & 0.2106 &  0.1563   & 0.177 \\ \midrule \midrule
        GPT-4V(Web)~\cite{openai2023gpt4v}    & 0.4 & -  &  -  & 0.586 &  -   & -  \\ \midrule
        GPT-4V(API)~\cite{openai2023gpt4v}  & 0.325 & 0.15  &  0.35  & 0.56 &  0.48   & 0.596  \\ \midrule

    \end{tabular}
    \caption{Main results in terms of accuracy. mPLUG-7B represents ``mPLUG-OWL-LLaMA''. miniGPT-7B represents ``miniGPT-4-Vicuna-7B''. CogVLM represents ``cogvlm-chat''. 
    GPT-4V(Web) represents the performance of GPT-4V on \tool\ with the prompt provided by the official web service. GPT-4V(API) is used by calling ``gpt-4-vision-preview.'' }
    \label{tab:result-acc}
\end{table*}

\subsection{Quantitative Analysis}
\label{subsec:quantitive-analysis}

\begin{figure}[!htpb]
    \centering
    \includegraphics[width=1.0\textwidth]{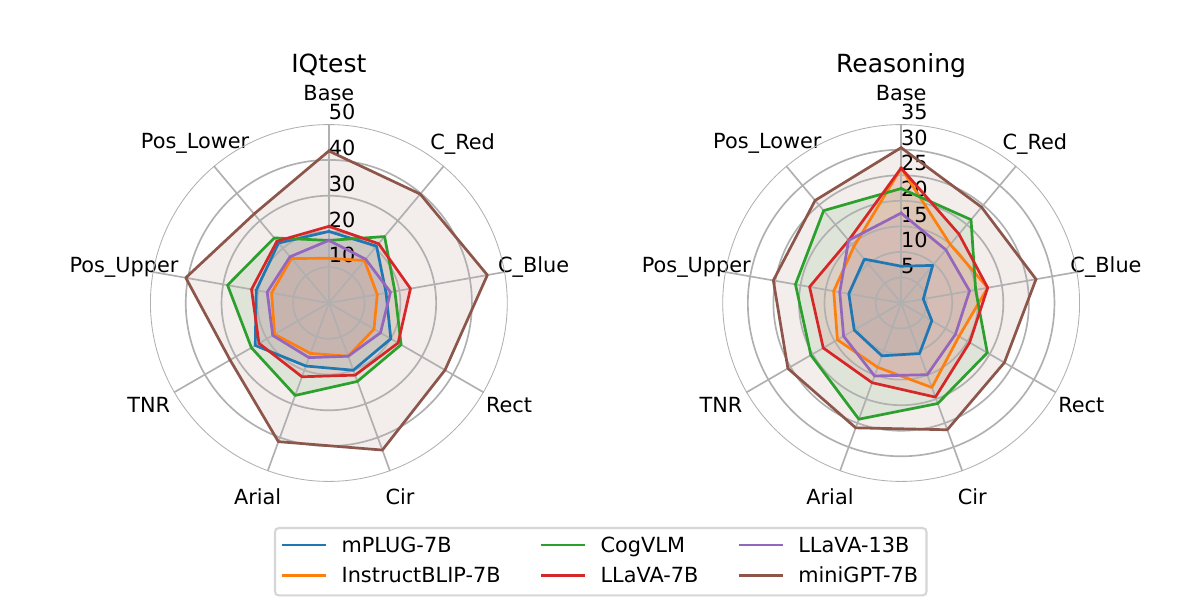} 
    \caption{Accuracies of six open-source LMMs on our proposed \tool\ across metamorphic relations.    
    \textbf{Left}: IQtest dataset, \textbf{Right}: Reasoning dataset.}
    \label{fig:vsp}
\end{figure}

To further understand the impact of different \vsp on \tool, we conduct a quantitative analysis in this section. 
The results are shown in~\F\ref{fig:vsp}, where two radar charts are provided for the IQtest and Reasoning datasets, respectively.
Besides the axis named ``Base'' (which represents the accuracy of the model without any intervention), we provide eight more 
axes represent interventions across 4 attributes: color (``C\_Red'', ``C\_Bleu''), position (``Pos\_Lower'', ``Pos\_Upper''), font (``Arial'', ``TNR''), and shape (``Cir'', ``Rect'').
The value of each axis represents the accuracy of the model with the corresponding intervention.

The radar charts facilitate analysis in two key ways. First, comparing an axis to its counterpart reveals the impact of a specific intervention. For instance, miniGPT-7B's lower ``Pos\_Lower'' value on IQtest indicates enhanced performance when the pointer is upper-positioned. Second, comparing an axis to ``Base'' shows the effect of that intervention. On the Reasoning dataset, mPLUG-7B has higher values of both ``TNR'' and ``Arial'' than ``Base'' on the Reasoning dataset, demonstrating that accuracy gets improved from font changes. 
Moreover, we can also cross-compare the performance of different LMMs and validate whether one specific LMM would possibly lose its leading position when a specific intervention is applied. Such scenarios occur in the radar chart as the lines of different colors cross. 
For example, on the Reasoning dataset, we observe that the red line (LLaVA-7B) and the green line (CogVLM) intersect between ``Base'' and ``Pos\_Lower,'' meaning LLaMA-7B outperforms without intervention, but CogVLM surpasses it with lower pointer positioning.
Such crossings highlight the importance of tailored \vsp\ selection in relatively boosting certain models. Notably, our experiments solely examine \vsp\ selection impacts, without considering other techniques like Chain-of-Thought~\cite{wei2022chain}; see~\S~\ref{sec:limitation} for further discussion.

\finding{2}{The choice of \vsp\ has a significant impact on the accuracy of LMMs. While the best \vsp\ combination varies for different LMMs and datasets, a targeted prompt selection allows an otherwise underperforming model to outperform other models.}

\subsection{Case Studies}
\label{subsec:case-study}

This section presents case studies that provide an intuitive understanding of the impact of \vsp, both positive and negative. Moreover, we provide some insight to avoid the negative impact of \vsp.
Due to space limitations, we do not include the images in this section, but interested readers can refer to~\cite{VisualReferPrompt} for more examples.

\parh{Positive Case.}~We find that \vsp\ can be beneficial in improving the performance of LMMs in cases where the original answer is incorrect. 
Specifically, we identify two types of incorrect answers: the first type occurs when the LMM provides a meaningful answer, but it is not the correct one, while the second type occurs when the LMM fails to provide a meaningful answer and simply repeats the question. In both scenarios, the visual reference can assist the LMM in identifying the intended meaning of the question, leading to the correct answer.
For example, in the 12th test case of the IQtest dataset, GPT-4V initially introduced the general rules for finding a missing image and further responded by asking, ``Would you prefer to solve the puzzle yourself?'' This response simply returns the question to the user. However, with the aid of Partial-intervention, GPT-4V is able to comprehend the question and provide the correct answer with detailed thinking steps. This illustrates the value of \vsp\ in improving the accuracy of LMMs in certain scenarios.

\parh{Negative Case.}~As shown in~\T~\ref{tab:result-acc}, GPT-4V
may perform worse with Partial-intervention on IQtest dataset than without any intervention.
Through manual analysis, it is discovered that GPT-4V's failure to provide a clear answer is a common reason for its failure to correctly respond to the Partial-interventioned variants, which had initially been answered correctly on the original image. Specifically, in over 78\% of cases, GPT-4V would request further information, such as ``please provide more information about the red box.'' Although this may be effective in real-world scenarios, in our test setting, it was considered a failure to answer the question as multi-turn interaction is out of our study scope.

\parh{Minimizing Negative Impact.}~In light of the potential negative impact of \vsp\ on LLMs' performance, it is important to consider how to maximize its benefits while minimizing its drawbacks. Fortunately, we identify two methods for avoiding the negative impact of \vsp\ in our experiments.
First, providing additional clarification can be used to mitigate the negative impact of \vsp. By giving more specific hints that describe the current situation, such as explicitly informing GPT-4V that an image with partial or Full-intervention will be provided and providing a further description of the definition of intervention, GPT-4V is less likely to refuse to respond.
Additionally, providing a vision reference with clear contrast and boundaries can also avoid the negative impact of \vsp. This is especially important when the tested images have multiple objects, and the questions are complex to answer. For instance, in the case of the 12th image of the Reasoning dataset, GPT-4V can correctly identify the missing picture when the vision reference shape is a circle but fails when the vision reference shape is replaced with a box. Although it may be apparent to humans that different reference shapes have the same intention in this scenario, this is not necessarily the case for GPT-4V. Therefore, providing a clear vision reference can help to avoid confusion and improve the accuracy of LLMs.

\section{Related Work}
\label{sec:related}

\parh{Prompting Scheme.}~Prompting enables LLMs and LMMs to perform specific tasks, representing a key component of their functionality. Traditionally, most models follow the paradigm of ``pre-train and fine-tune,'' where large-scale pre-training on corpora is followed by task-specific fine-tuning on limited data. However, the fine-tuning process is time-consuming and requires a large amount of data, which is not always available in real-world scenarios. 
To address this issue, recent LLMs have adopted a new paradigm called ``pre-train and prompt'' to enable zero-shot or few-shot learning without extensive fine-tuning. Through prompting, models can be directly applied to novel tasks using few or no examples, sidestepping data limitations. Prompting maintains efficacy for multimodal LMMs as well~\cite{yang2023dawn}.

\parh{Benchmarks for LMMs.}~ The assessment of large language/vision models often relies on standardized benchmarks meant to evaluate performance across diverse tasks. Existing benchmarks for LLMs generally fall into two categories: those focused on general language proficiency, like AlpacaEval~\cite{alpacaeval}, OpenLLM~\cite{leaderboard}, and Chatbot Arena~\cite{chatbotarena}, and those tailored to specific domains, such as medical question answering~\cite{singhal2022large}, emotion recognition~\cite{huang2023emotionally}, and ethical reasoning~\cite{huang2023trustgpt}. Similarly, benchmarks for LMMs include both multimodal datasets for tasks like visual question answering~\cite{antol2015vqa}, natural language visualization reasoning~\cite{suhr2017corpus}, and referring expressions~\cite{yu2016modeling}, as well as newer benchmarks probing advanced reasoning skills, like mathematical reasoning~\cite{lu2023mathvista} and multi-turn instruction following~\cite{guo2023pptc}. These benchmarks allow standardized comparison of different models' capabilities across diverse tasks.

\parh{Metamorphic Testing.}~Metamorphic testing (MT) leverages metamorphic relations between inputs to uncover potential errors in systems under test~\cite{chen1998metamorphic}. By generating new inputs and outputs based on predefined relations, MT can effectively reveal bugs across domains like autonomous driving~\cite{zhang2018deeproad} and object detection~\cite{wang2020metamorphic}. Recently, MT has been applied to evaluate large language models. For instance, CCTEST~\cite{li2022cctest} uses MT to synthesize code variants and uncover inconsistencies in code completion systems like GitHub Copilot~\cite{copilot}. Additionally, metamorphic relations have been designed to find hundreds of logical errors in machine translation services~\cite{sun2020automatic} like Google Translate~\cite{googletrans}. For vision tasks, MT has been used to evaluate the robustness of visual question answering systems~\cite{yuan2021perception} and uncover biases in images. In this work, we leverage MT to systematically generate new VRP combinations and evaluate the performance of LMMs on \vsp\ tasks.

\section{Limitations}
\label{sec:limitation}

While we have evaluated eight LMMs using the \tool benchmark in \S~\ref{sec:method}, our evaluation and dataset are still in the preliminary stages and could prompt further research from the following perspectives:

\parh{LMM Maturity.}~While LMMs represent an exciting new frontier in AI, this study acknowledges that such models are still in their nascent stages and have room for improvement. As reported by OpenAI~\cite{openai2023gpt4v}, even advanced LMMs like GPT-4V can fail on seemingly simple tasks or be vulnerable to adversarial attacks, indicating their performance is far from perfect. For example, it may give the opposite answer when the tested image slightly rotates~\cite{liu2023hallusionbench}. However, we believe the potential of LMMs for multimodal scenarios merits exploration, and our study aims to offer initial insights to guide future research. As LMMs mature, their capabilities for \vsp\ will likely grow. But for now, they remain imperfect tools whose limitations must be recognized, even as we investigate their possibilities. We hope our work spurs additional progress in this burgeoning field.

\parh{Task Coverage.}~The dataset used to benchmark LMMs' performance has limited diversity and scale, which restricts the breadth of capabilities we can evaluate. Expanding to larger, more varied tasks could better assess performance across different contexts. However, some tasks lack unambiguous ground truth answers (e.g. explaining an image romantically), making it challenging to define automatic evaluation metrics that apply universally. Overall, while not exhaustive, we believe the current benchmark sufficiently evaluates LMM performance on \vsp\ and provides a valuable baseline for future work. As datasets grow more diverse, \vsp\ capabilities on a wider range of multimodal tasks can be tested.

\parh{Few-shot Learning and Chain-of-Thought.}~In this work, we only consider zero-shot~\cite{liu2021pre} prompting strategy, where only the task definition is provided to the model. However, performance may improve by supplying additional in-context examples~\cite{wei2022emergent}. Among a series of works in prompt engineering proposed~\cite{wei2022chain, wang2022self, sanh2021multitask, DBLP:conf/emnlp/ShinRLWS20}, the chain-of-thought (CoT) approach shows particular promise, requiring models to provide intermediate reasoning steps (rationales) before the final output~\cite{wei2022chain,wang2022rationale,li2022explanations,li2022advance}.
CoT has been shown to be effective in improving reasoning abilities in LLMs.
Notably, current few-shot learning and CoT strategies are still limited to text-based prompting, and it is still unclear how to extend them to visual referring prompting. We believe that this is an interesting direction for future research.

\section{Conclusion}
\label{sec:conclusion}

In this work, we introduced a benchmark and conducted an initial comprehensive study evaluating LMMs in diverse visual referring prompting settings. Through a manually curated dataset and MT-based framework, we assessed the performance of prominent open-source and proprietary foundation models, including several early versions of GPT-4V. 
Our study revealed intriguing preliminary findings into the impact of visual referring prompts, while also exploring GPT-4V's capabilities. 
This research laid the groundwork to advance understanding of LMMs' capabilities under \vsp. 
While not exhaustive, our benchmark provided an important first step toward standardized assessment. 
We hope this pioneering study stimulates further research into how visual context interacts with LMMs. By establishing initial baselines, our work aims to catalyze continued progress in this burgeoning area.

\bibliography{bib/cv,bib/llm,bib/cot,bib/sast,bib/others,bib/zj,bib/lmm,bib/ref,bib/analysis,bib/decompiler,bib/testing-cv,bib/reference}
\bibliographystyle{plain}


\end{document}